\documentclass{article}
\usepackage{spconf,amsmath,graphicx}
\usepackage{makecell}
\usepackage{multirow}
\usepackage{multicol}
\usepackage{subfigure}
\usepackage{verbatim}
\usepackage{amsmath}
\usepackage{amssymb}
\usepackage{pifont}
\usepackage{color}
\usepackage{xcolor}
\usepackage{comment}
\newcommand{\cmark}{\ding{51}}%
\newcommand{\figref}[1]{Fig. \ref{#1}}
\newcommand{\tabref}[1]{Table \ref{#1}}

\makeatletter
\def\hlinewd#1{%
\noalign{\ifnum0=`}\fi\hrule \@height #1 \futurelet
\reserved@a\@xhline}

\title{Joint Learning of Feature Extraction and Cost Aggregation for Semantic Correspondence}
\name{Jiwon Kim, Youngjo Min, Mira Kim, and Seungryong Kim}
\address{Computer Vision Lab. (CVLAB), Korea University, Korea}

\begin{document}
%
\maketitle
\begin{abstract}
Establishing dense correspondences across semantically similar images is one of the challenging tasks due to the significant intra-class variations and background clutters. To solve these problems, numerous methods have been proposed, focused on learning feature extractor or cost aggregation independently, which yields sub-optimal performance. In this paper, we propose a novel framework for jointly learning feature extraction and cost aggregation for semantic correspondence. By exploiting the pseudo labels from each module, the networks consisting of feature extraction and cost aggregation modules are simultaneously learned in a boosting fashion. Moreover, to ignore unreliable pseudo labels, we present a confidence-aware contrastive loss function for learning the networks in a weakly-supervised manner. We demonstrate our competitive results on standard benchmarks for semantic correspondence.
\end{abstract}
\section{Introduction}
\label{sec:intro}
Semantic correspondence is one of the essential tasks on various Computer Vision applications~\cite{taira2019right, wang2020learning, cho2021semantic}, which generally aims to establish pixel-wise, but locally-consistent correspondences across semantically similar images. It is an extremely challenging task because finding semantic correspondences can be easily distracted by non-rigid deformations and large variations on the appearance within the same class~\cite{lee2019sfnet}.

Recent methods~\cite{rocco2018neighbourhood,cho2021semantic} solved the task by designing deep Convolutional Neural Networks (CNNs). The networks often consist of \textit{feature extraction} and \textit{cost aggregation} steps. 
For the feature extraction step, instead of relying on hand-crafted descriptors as in conventional methods~\cite{lowe2004distinctive}, recently, there has been an increasing interest in leveraging the representation power of CNNs~\cite{choy2016universal, kim2017fcss,lee2019sfnet}. However, they can struggle with determining the correct matches from the cost volume because ambiguous matching pairs are often generated by repetitive patterns and occlusions.
For cost aggregation step, methods~\cite{rocco2017convolutional, rocco2018end, rocco2018neighbourhood, melekhov2019dgc, rocco2020efficient, li2020correspondence, lee2021patchmatch, cho2021semantic} attempted to determine the correct matches between the great majority of dense information and non-distinctive matching pairs. Unlike previous strategies~\cite{lowe2004distinctive,leibe2008robust}, recent methods proposed a trainable matching cost aggregation in the overall network~\cite{rocco2017convolutional, rocco2018end, rocco2018neighbourhood, melekhov2019dgc, rocco2020efficient, li2020correspondence, lee2021patchmatch, cho2021semantic}.

Learning semantic correspondence networks consisting of feature extraction and cost aggregation modules in a \textit{supervised} manner requires a large-scale ground-truth which is notoriously hard to build. To alleviate this, several methods leveraged \textit{pseudo-labels}, extracted from networks' prediction itself by Winner-Take-All (WTA), and train the networks in an unsupervised\footnote{They are often also called weak-supervised methods since they require the image pairs.} manner~\cite{rocco2018end, rocco2018neighbourhood}. Although they are appealing alternatives, they are sensitive to uncertain pseudo labels. 
Furthermore, jointly using the pseudo labels from feature extraction and cost aggregation modules may boost the performance, but there was no study for this approach.

Some recent self-supervised methods~\cite{park2020contrastive, wang2020dense} use dense contrastive loss for pixel-wise prediction tasks instead of using image-level contrastive loss. However, they do not consider unconfident matches generated from repetitive fields or occlusions and semantic appearance variations. 
\begin{figure*}[t]
	\centering
	\renewcommand{\thesubfigure}{}
	\subfigure[(a) Learning feature extraction only]
	{\includegraphics[width=0.3\linewidth]{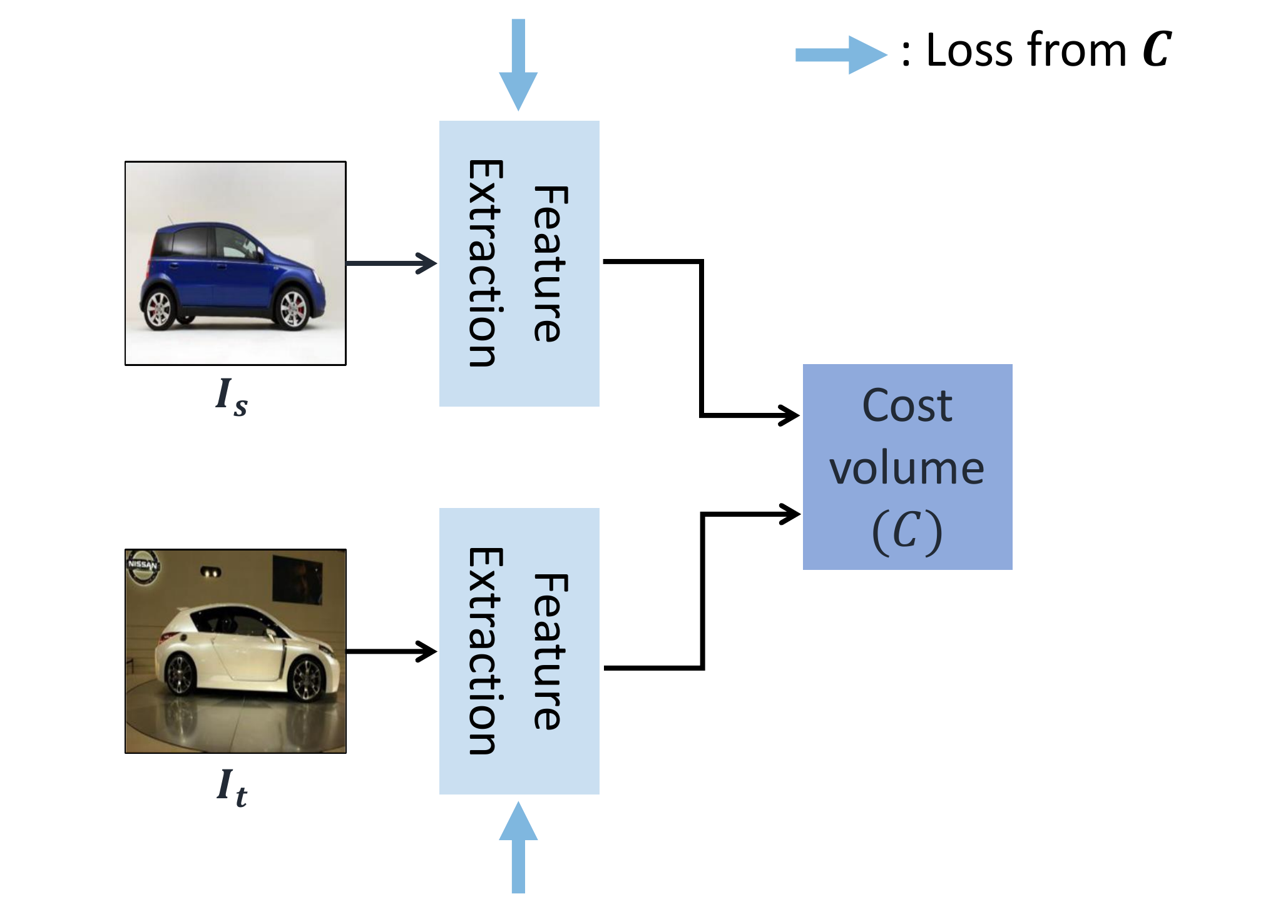}}\hfill
	\subfigure[(b) Learning cost aggregation only]
	{\includegraphics[width=0.3\linewidth]{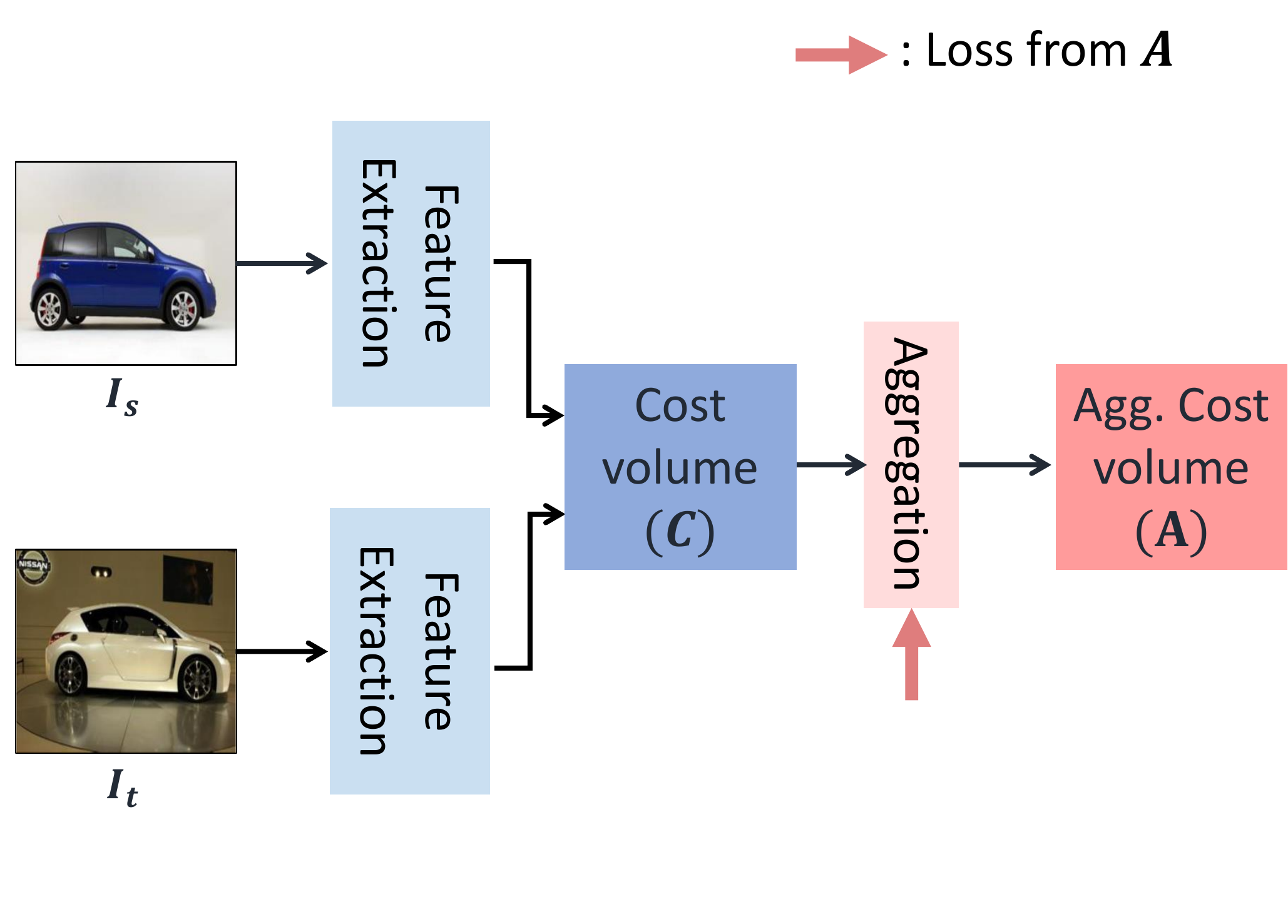}}\hfill
	\subfigure[(c) Ours]
	{\includegraphics[width=0.3\linewidth]{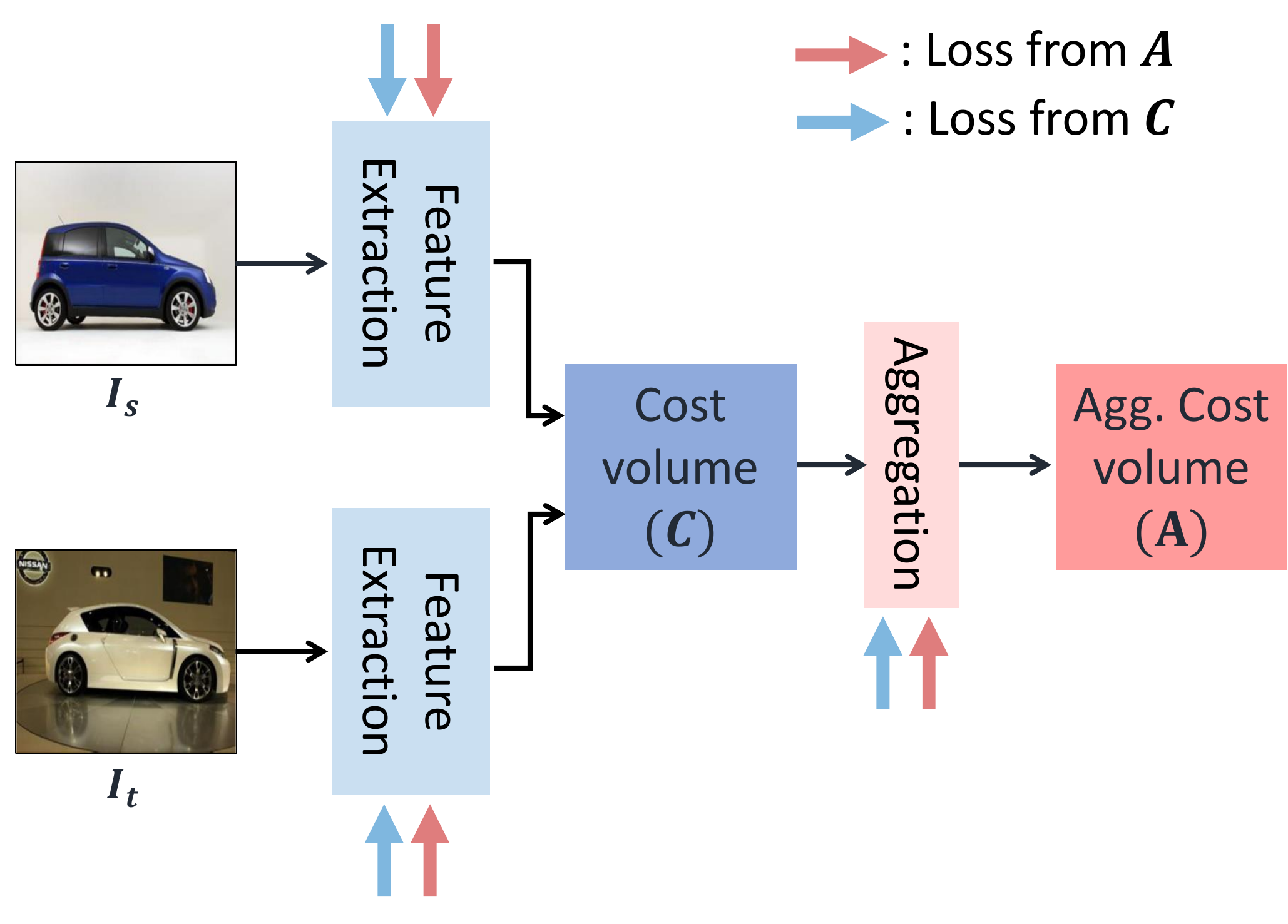}}\hfill\\
	\vspace{-10pt}
	\caption{\textbf{Intuition of our method:} (a) methods for solely training a feature extraction network~\cite{choy2016universal, kim2017fcss, lee2019sfnet}, (b) methods for solely training a cost aggregation network for filtering the ambiguous matching cost~\cite{rocco2017convolutional, rocco2018end, rocco2018neighbourhood, melekhov2019dgc, lee2021patchmatch}, and (c) Ours, which jointly training feature extraction and cost aggregation networks with the proposed loss function.}
	\label{fig:problem_statement}
	\vspace{-10pt}    
\end{figure*}

In this work, we present a novel framework that jointly learns feature extraction and cost aggregation for semantic correspondence. Motivated by the observation that the pseudo labels from feature extraction and cost aggregation steps can complement each other, we encourage the feature extraction and cost aggregation modules to be jointly trained by exploiting the pseudo-labels from each other module. In addition, to filter out unreliable pseudo-labels, we present a confidence-aware contrastive loss function that exploits a forward-backward consistency to remove incorrect matches. We demonstrate experiments on several benchmarks~\cite{ham2017proposal,taniai2016joint}, proving the robustness of the proposed method over the latest methods for semantic correspondence.

\section{Methodology}
\label{sec:methodology}
\subsection{Preliminaries}
Given a pair of images, i.e., source $I_{s}$ and target $I_{t}$, which represent semantically similar images, the objective of semantic correspondence is to predict dense correspondence fields $P$ between the two images at each pixel. To achieve this, most predominant methods consist of two steps, namely \textit{feature extraction} and \textit{cost aggregation}~\cite{kim2017fcss, rocco2018neighbourhood}. In specific, the first step is to apply feature extraction networks to obtain a 3D tensor $D \in \mathbb{R}^{h \times w \times d}$, where $h \times w$ is the spatial resolution of the feature maps and $d$ denotes the number of channels. To estimate correspondences, the similarities between feature maps $
D_s$ and $D_t$ from source and target images, respectively, are measured, which outputs a 4D cost volume $\mathcal{C}(i, j)$, where ${i, j \in \{1,...,h \times w\}}$, through a cosine similarity score such that $\mathcal{C}(i, j)=D_s(i)^TD_t(j)$. Estimating the correspondence with sole reliance on matching similarities is sensitive to matching outliers, and thus the cost aggregation steps are used to refine the initial matching similarities to achieve the aggregated cost $\mathcal{A}(i, j)$ through cost aggregation networks.

Learning such networks, i.e., feature extraction and cost aggregation modules, in a \textit{supervised} manner requires manually annotated ground-truth correspondences $P^*$, which is extremely labor-intensive and subjective to build~\cite{taniai2016joint,ham2017proposal,rocco2018neighbourhood}. To overcome this challenge, an alternative way is to leverage Winner-Take-All (WTA) matching point, which is the most likely match by an argmax function on $\mathcal{C}(i,j)$ or $\mathcal{A}(i,j)$, as a pseudo correspondence label $F$. For instance, NCNet~\cite{rocco2018neighbourhood} (and its variants~\cite{li2020correspondence, rocco2020efficient, lee2021patchmatch}) and DenseCL~\cite{wang2020dense} utilized such correspondences $F$ to learn the cost aggregation networks and feature extraction networks, respectively, in an \textit{unsupervised} fashion, as exemplified in~\figref{fig:problem_statement}. Although they are definitely appealing alternatives, these frameworks are highly sensitive to \textit{uncertain} pseudo labels. Moreover, there exists no study to jointly train the feature extraction networks and cost aggregation networks in a complementary and boosting manner. \vspace{-10pt}  

\begin{figure*}[t]
 	\centering
 	\renewcommand{\thesubfigure}{}
 	\subfigure[(a) $I_s$]
 	{\includegraphics[width=0.122\linewidth]{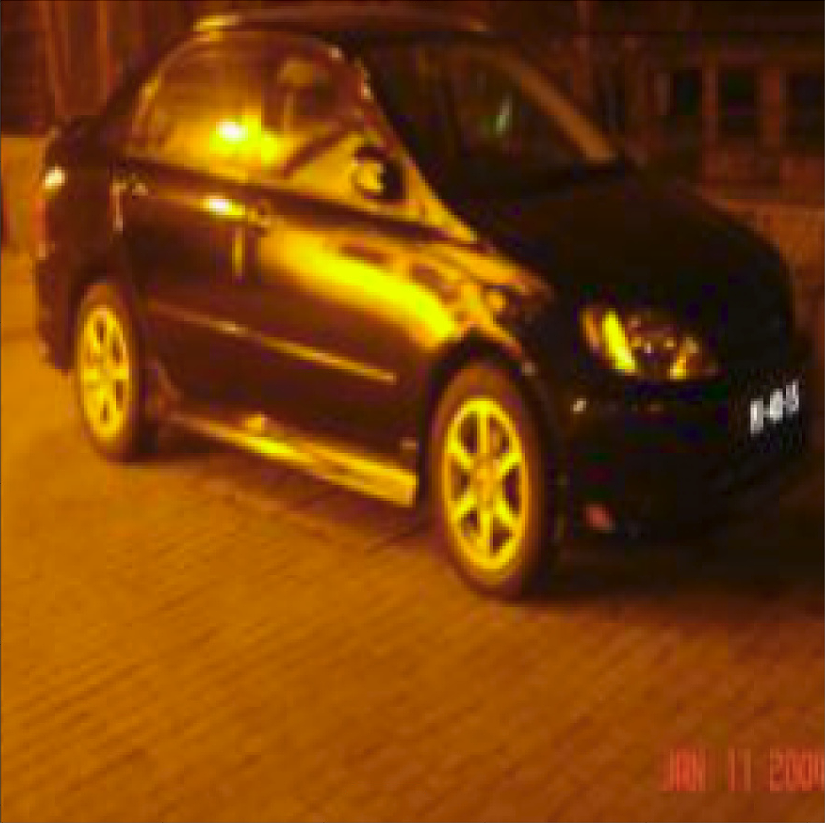}}\hfill 	
  	\subfigure[(b) $I_t$]
  	{\includegraphics[width=0.122\linewidth]{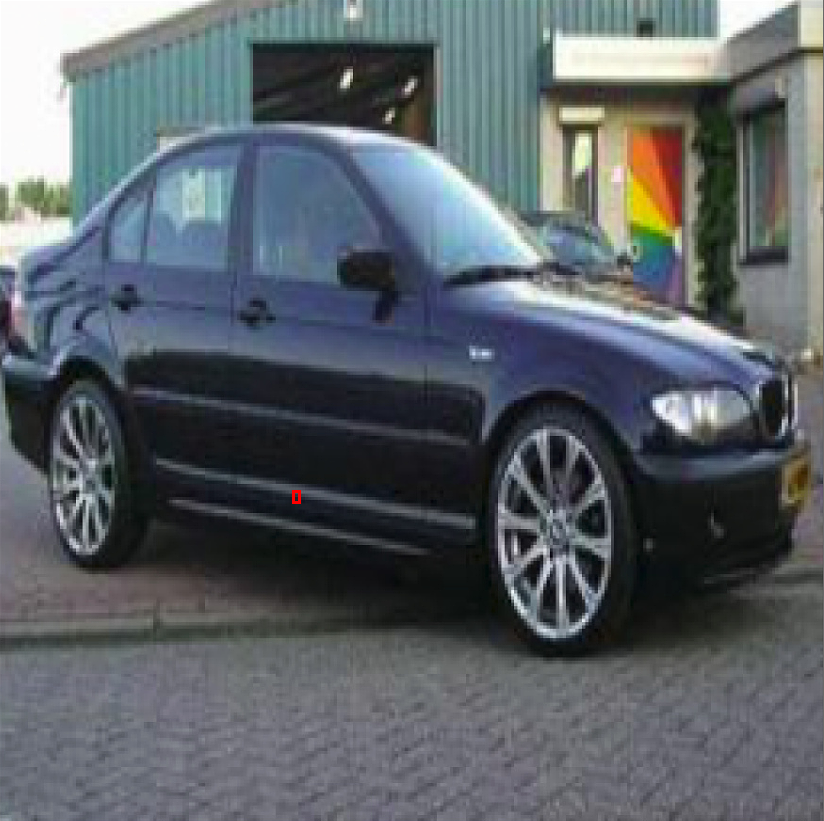}}\hfill
	\subfigure[(c) $F^{\mathcal{C},0}_{s \rightarrow t}$]
	{\includegraphics[width=0.122\linewidth]{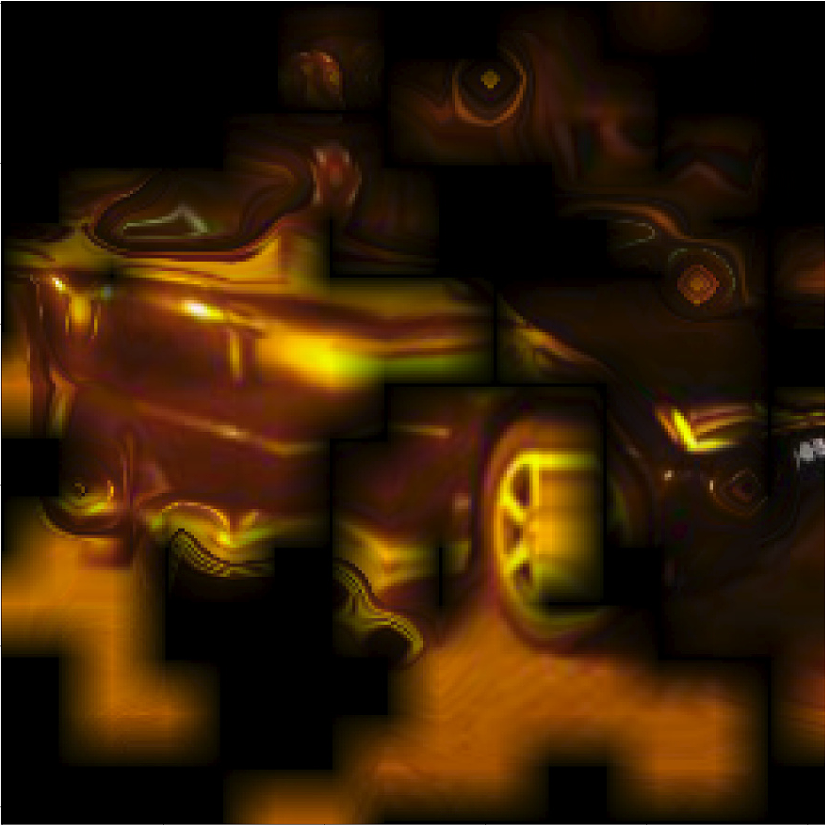}}\hfill
	\subfigure[(d) $F^{\mathcal{A},0}_{s \rightarrow t}$]
 	{\includegraphics[width=0.122\linewidth]{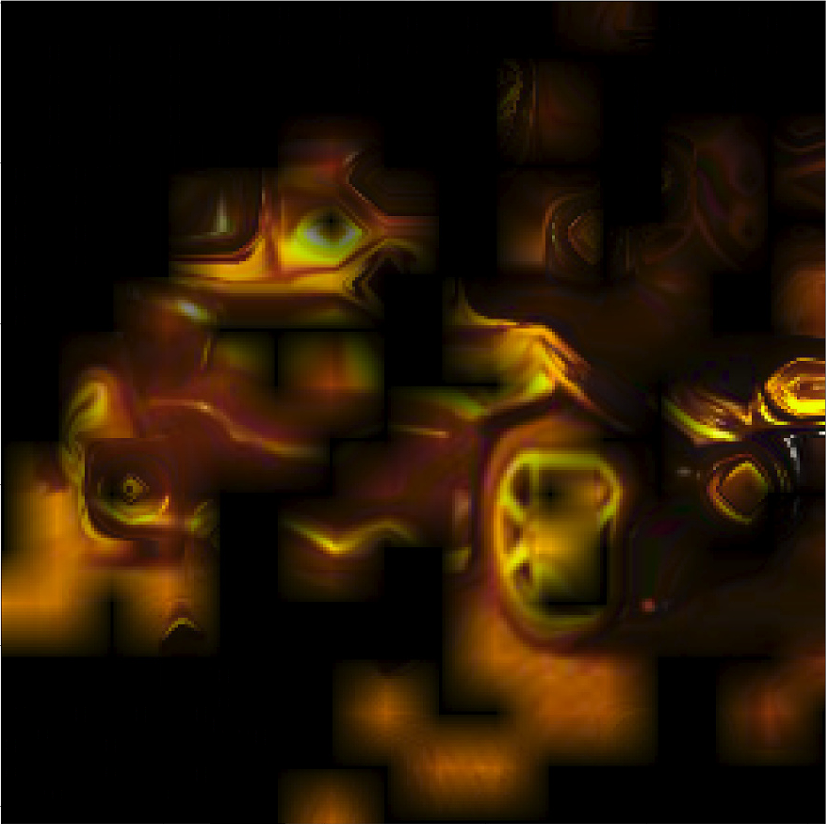}}\hfill
 	\subfigure[(e) $F^{\mathcal{C},500}_{s \rightarrow t}$]
	{\includegraphics[width=0.122\linewidth]{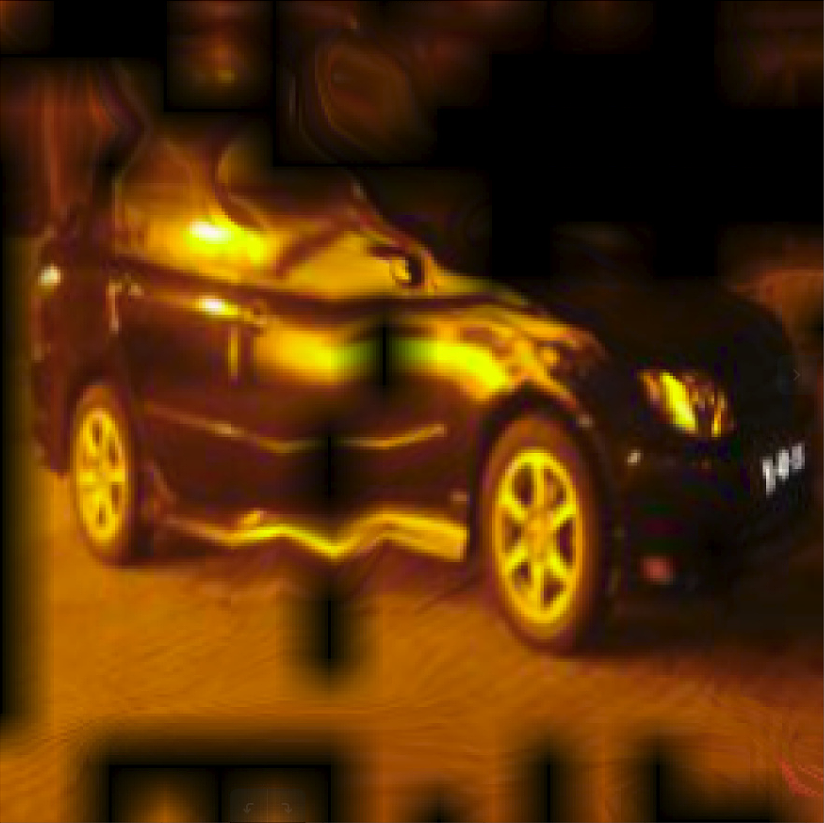}}\hfill
 	\subfigure[(f) $F^{\mathcal{A},500}_{s \rightarrow t}$]
	{\includegraphics[width=0.122\linewidth]{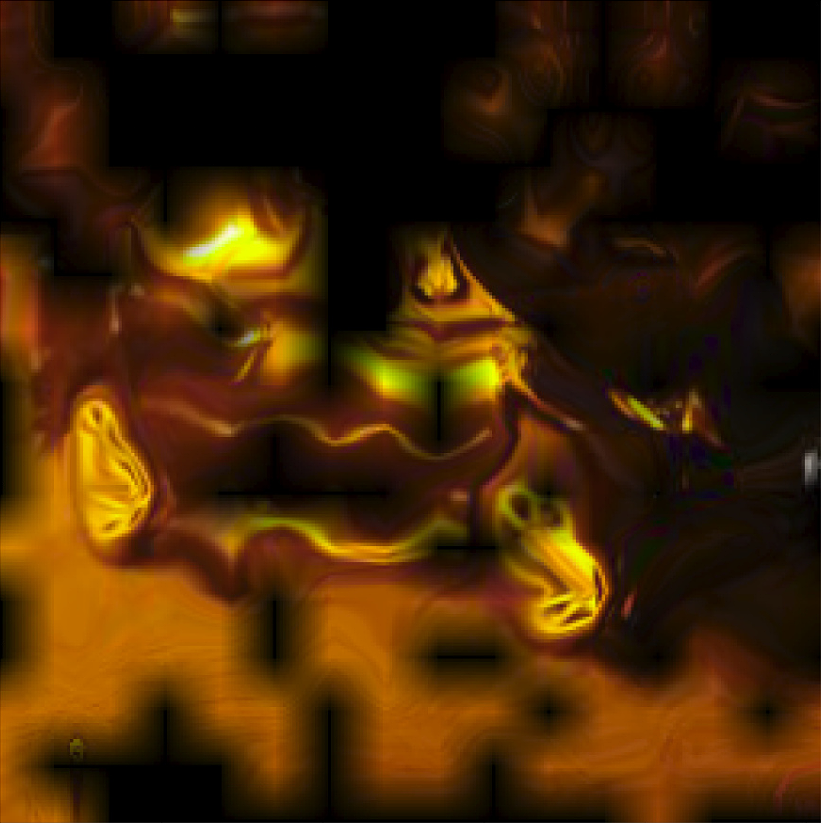}}\hfill
 	\subfigure[(g) $F^{\mathcal{C},1000}_{s \rightarrow t}$]
	{\includegraphics[width=0.122\linewidth]{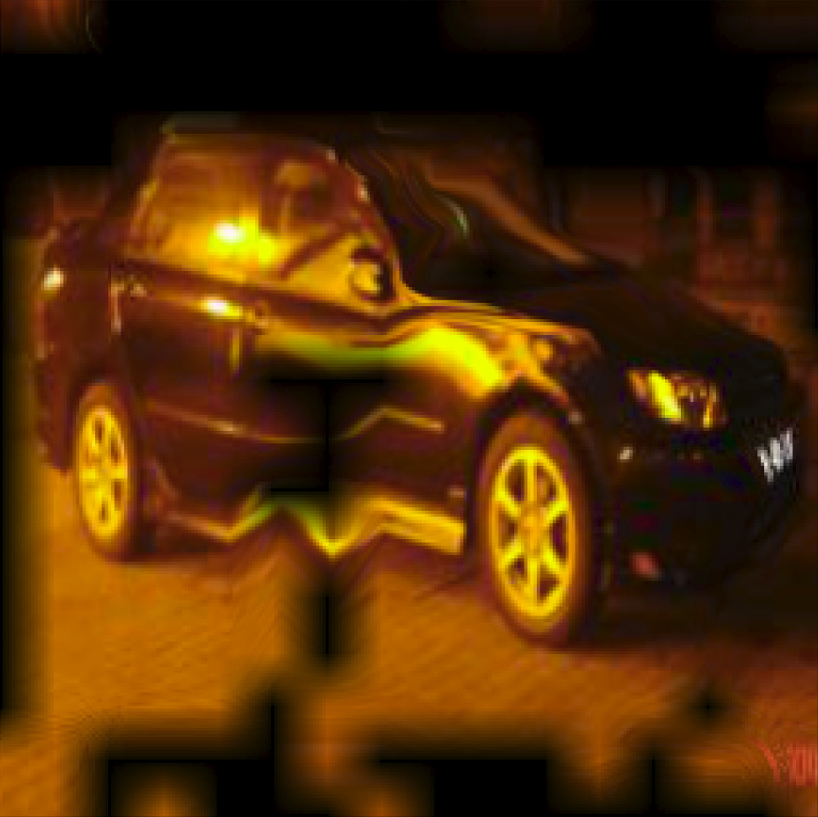}}\hfill
 	\subfigure[(h) $F^{\mathcal{A},1000}_{s \rightarrow t}$]
	{\includegraphics[width=0.122\linewidth]{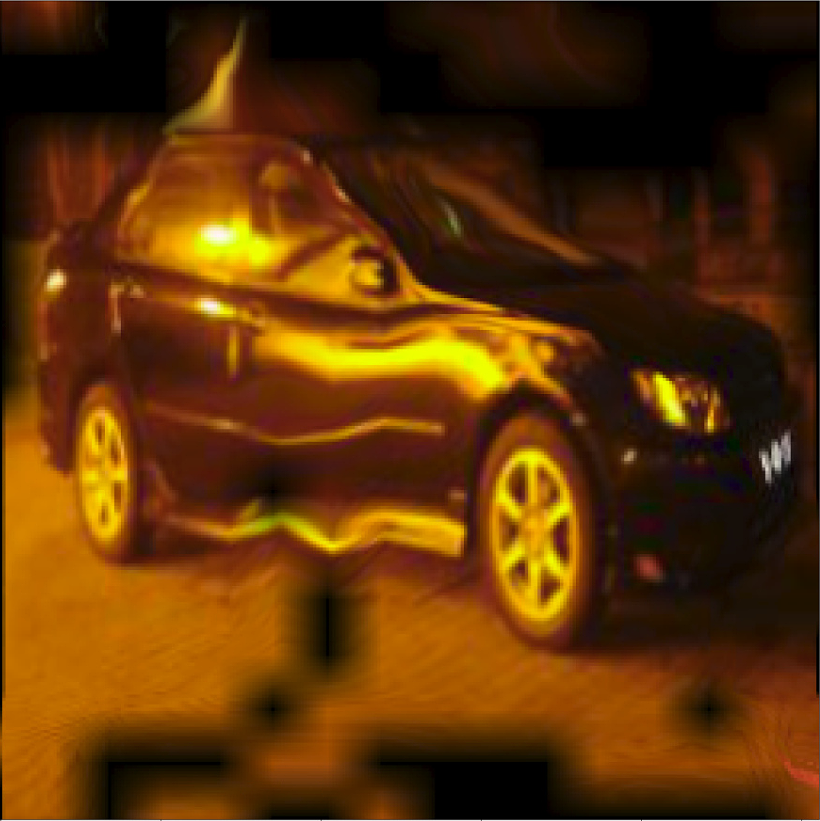}}\hfill
 	\\\vspace{-10pt}
 	\caption{\textbf{Comparison on PF-Pascal~\cite{ham2017proposal}  applying masking and warping by the predicted correspondence map.} At later iteration (1000) with $F^{\mathcal{A},1000}_{s \rightarrow t}$ have more confident correspondences than with $F^{\mathcal{C},1000}_{s \rightarrow t}$ and at earlier iteration (500) vice versa.}\label{fig:ablation_iter}\vspace{-10pt}
\end{figure*}


\subsection{Confidence-aware Contrastive Learning}
In this section, we first study how to achieve better pseudo labels for dense correspondence, and then present a confidence-aware contrastive learning.

We start from classic uncertainty measurement based on forward-backward consistency checking as proposed in~\cite{sundaram2010dense,meister2018unflow, liu2019ddflow}, where an argmax operator was applied twice for forward and backward directions, respectively. Specifically, the pseudo matching map $F^\mathcal{C}_{s \rightarrow t}$ from the matching cost $\mathcal{C}$, warping $I_s$ toward $I_t$, is defined as follows:
\begin{equation}
F^\mathcal{C}_{s \rightarrow t}(i)=\mathrm{argmax}_{j'} \,\mathcal{C}\left(i,j'\right)  -i
\end{equation}
where $j'$ is defined for all the points in the target. Similarly, $F^\mathcal{C}_{t \rightarrow s}$ can be computed, warping $I_t$ toward $I_s$. In an non-occlusion region, we get a backward flow vector $F^\mathcal{C}_{t \rightarrow s}$ in the inverse direction as the forward flow vector $F^\mathcal{C}_{s \rightarrow t}$. If this consistency constraint is not satisfied, the points in the target are occluded at the matches in the source, or the estimated flow vector is incorrect. These constraints can be defined such that 
\begin{equation}
\begin{split}
\|F^\mathcal{C}_{s \rightarrow t}(i)+&F^\mathcal{C}_{t \rightarrow s}(i+F^\mathcal{C}_{s \rightarrow t}(i))\|^{2} \\
<
&{\alpha}_1(\|F^\mathcal{C}_{s \rightarrow t}(i)\|^{2}+\|F^\mathcal{C}_{t \rightarrow s}(i+F^\mathcal{C}_{s \rightarrow t}(i))\|^{2})+{\alpha}_2.
\end{split}
\end{equation}
Since there may be some estimation errors in the flows, we grant a tolerance interval by setting hyper-parameters ${\alpha}_1$ and ${\alpha}_2$. 
A binary mask $M^\mathcal{C}$ is then obtained by such forward-backward consistency checking, representing a non-occluded region as 1 and an occluded region as 0.

Based on the estimated mask $M^\mathcal{C}$, we present a confidence-aware contrastive loss function, aiming to maximize
the similarities at the reliably matched points while minimizing for
the others, defined such that
\begin{equation}
    \mathcal{L}^\mathcal{C}_\mathrm{ccl} = - \frac{1}{N^\mathcal{C}}\sum_i M^\mathcal{C}(i) \mathrm{log}\left(\frac{\mathrm{exp}(\mathcal{C}(i,i+F^\mathcal{C}_{s \rightarrow t}(i))/\gamma)}{\sum_j \mathrm{exp}(\mathcal{C}(i,j)/\gamma)}\right),
\end{equation}
where $N^\mathcal{C}$ is the number of non-occluded pixels, and $\gamma$ is a temperature hyper-parameter. Our loss function enables rejecting ambiguous matches with a thresholding while
accepting the confident matches. While $\mathcal{L}^\mathcal{C}_\mathrm{ccl}$ is formulated to train the feature extraction network itself, this loss function can also be defined for aggregated cost $\mathcal{A}$ as $\mathcal{L}^\mathcal{A}_\mathrm{ccl}$, which can be used to train the cost aggregation networks. 
\begin{figure}[t]
\centering
    \includegraphics[width=0.9\linewidth]{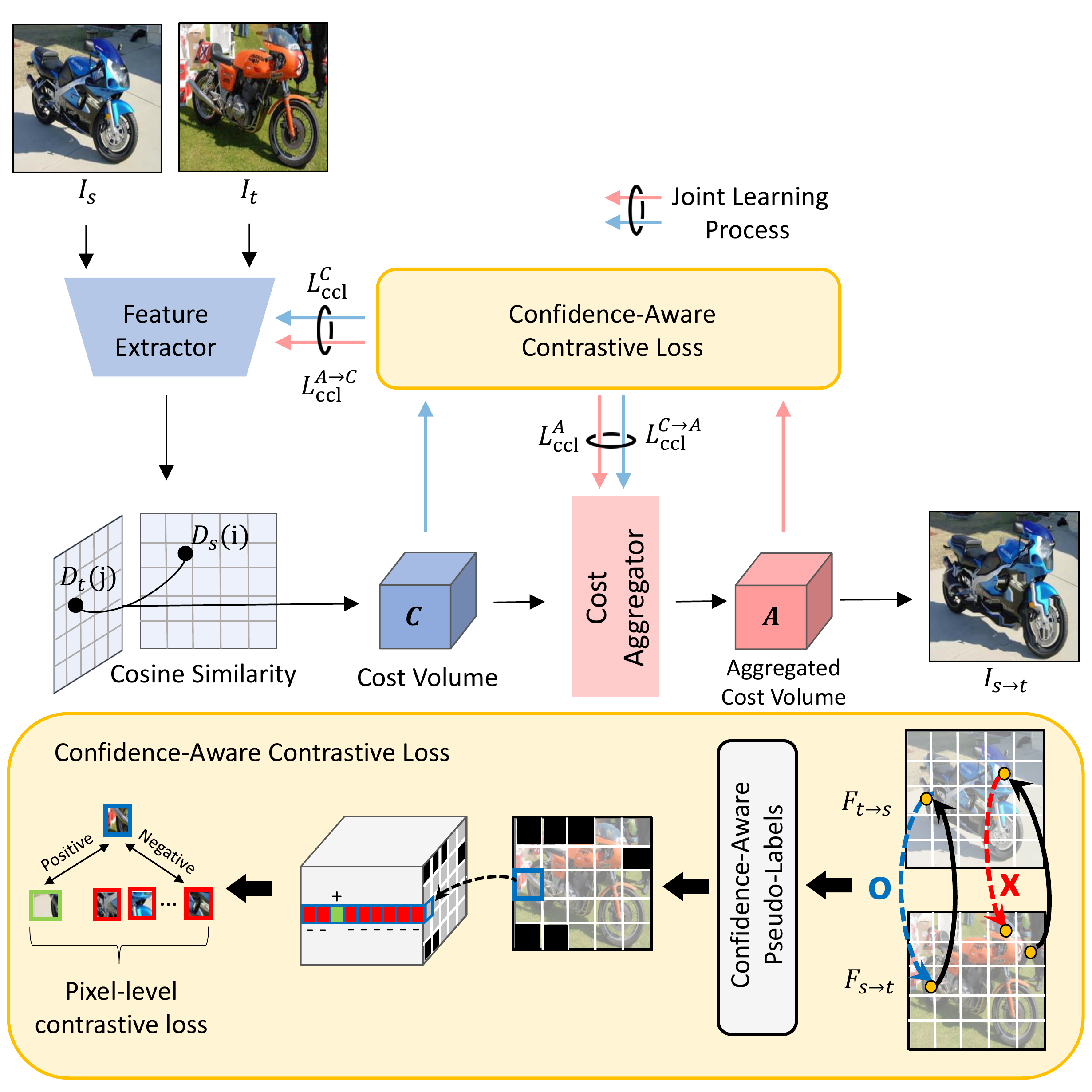}\hfill \\ 
    \caption{\textbf{Overall framework.}} 
    \label{fig:net_architecture}\vspace{-10pt} 
\end{figure}

\subsection{Joint Learning}
The proposed confidence-aware contrastive loss function can be independently used for learning feature extraction and cost aggregation networks through $\mathcal{L}^\mathcal{C}_\mathrm{ccl}$ and $\mathcal{L}^\mathcal{A}_\mathrm{ccl}$, respectively. However, since two pseudo labels from feature extraction networks and cost aggregation networks may have complementary information, using two pseudo labels in a joint manner can enable further boosting the performance. For instance, at early stages of training, the pseudo label by \textit{pre-trained} feature extractor provides more reliable cues than ones by \textit{randomly-initialized} cost aggregation networks, which may help the cost aggregation networks converge much faster, as exemplified in~\figref{fig:ablation_iter}. In addition, as the training progresses, the \textit{well-trained} cost aggregation networks produce superior correspondences than the pseudo label by feature extractor, as exemplified in~\figref{fig:ablation_iter}.  
\begin{table*}[t!] 
\centering
\scalebox{0.8}
{
\begin{tabular}{c|c|c|c|c|c|c|cccc}

\hlinewd{0.8pt}
\multirow{2}{*}{Method} & \multirow{2}{*}{Base} & \multicolumn{2}{c|}{Trainable Comp.} & \multirow{2}{*}{Joint} & \multirow{2}{*}{PF-PSCAL ($\alpha$ = 0.1)} & \multirow{2}{*}{PF-Willow ($\alpha$ = 0.1)} & \multicolumn{4}{c}{TSS ($\alpha$ = 0.05)}    \\ \cline{3-4} \cline{8-11} 
&   & Feature & Aggreg.                    &  &   &  & FG3D & JODS & PASCAL & Avg \\ \hline
WTA                                 &ResNet-101 &- & - & - & 53.3 & 46.9 &-&-&-&-\\ \hline
WeakAlign~\cite{rocco2018end}       &ResNet-101 &\cmark & \cmark &- & 75.8      & 84.3      & 
90.3 & 76.4 & 56.5  & 74.4 \\
RTNs~\cite{kim2018recurrent}         &ResNet-101 &- & \cmark &-& 75.9      & 71.9      & 
90.1 & 78.2 & 63.3  & 72.2 \\
NCNet~\cite{rocco2018neighbourhood} &ResNet-101 &- & \cmark &-& 78.9      & 84.3      &
94.5  & 81.4 & 57.1  & 77.7 \\
DCCNet~\cite{huang2019dynamic}      &ResNet-101 &\cmark & \cmark &-& 82.3      &73.8       &
93.5  &\textbf{82.6}  &57.6   &77.9  \\ 
ANCNet~\cite{li2020correspondence} &ResNet-101 &\cmark &\cmark & - & 86.1 & - & 
- & - & -& - \\
PMNC~\cite{lee2021patchmatch} & ResNet-101 & - &\cmark &- & 90.6 & - &
- & - & - & - \\
CATs~\cite{cho2021semantic}        &ResNet-101 &\cmark & \cmark  &- & 87.3    &76.9       &85.3      &73.7      &55.4      &73.6  \\
\hline
\textbf{Ours w/NCNet}                           &ResNet-101 &\cmark &\cmark &\cmark& 80.0     & \textbf{86.6}&\textbf{95.0} & 82.3 &55.8 & 78.4     \\ 
\textbf{Ours w/CATs}                            &ResNet-101 &\cmark & \cmark &\cmark& \textbf{92.5}    & 79.8&91.7 & 81.2 &~\textbf{60.9} & ~\textbf{80.0}  \\ 
\hlinewd{0.8pt}
\end{tabular}
}
\vspace{-5pt}
\caption{\textbf{Comparison with state-of-the-art methods on standard benchmarks~\cite{ham2017proposal,taniai2016joint}.}}
\vspace{-10pt}
\label{tab:quan}
\end{table*}

To leverage complementary information during training, we use a pseudo label output of each module, namely feature extraction and cost aggregation modules, defined such that
\begin{equation}
\begin{split}
    &\mathcal{L}^{\mathcal{A} \rightarrow \mathcal{C}}_\mathrm{ccl} = \\
    &- \frac{1}{N^\mathcal{A}}\sum_i M^\mathcal{A}(i) \mathrm{log}\left(\frac{\mathrm{exp}(\mathcal{C}(i,i+F^\mathcal{A}_{s \rightarrow t}(i))/\gamma)}{\sum_j \mathrm{exp}(\mathcal{C}(i,j)/\gamma)}\right),
\end{split}
\end{equation}
where $
F^\mathcal{A}_{s \rightarrow t}(i)={\mathrm{argmax}_{j'}}  \,\mathcal{A}\left(i,j'\right)$. $\mathcal{L}^{\mathcal{C} \rightarrow \mathcal{A}}_\mathrm{ccl}$ is similarly defined. Our final loss function is thus defined such that
\begin{equation}
    \mathcal{L} = \lambda^\mathcal{C} ( \mathcal{L}^{\mathcal{C}}_\mathrm{ccl} + \mathcal{L}^{\mathcal{A} \rightarrow \mathcal{C}}_\mathrm{ccl} ) + \lambda^\mathcal{A} ( \mathcal{L}^{\mathcal{A}}_\mathrm{ccl} + \mathcal{L}^{\mathcal{C} \rightarrow \mathcal{A}}_\mathrm{ccl} ), 
\end{equation}
where $\lambda^\mathcal{C}$ and $\lambda^\mathcal{A}$ represent hyper-parameters. 
\figref{fig:net_architecture} illustrates the overall architecture of the proposed methods.

\section{Experiments}
\subsection{Implementation Details}
In our framework, we used the ResNet-101~\cite{he2016deep} pretrained on ImageNet~\cite{deng2009imagenet} benchmark. We added additional layers followed by this to transform features to be highly discriminative w.r.t. both appearance and spatial context~\cite{lee2019sfnet}. 
In addition, we used two types of cost aggregation modules like 4D CNN~\cite{rocco2018neighbourhood}, denoted Ours w/NCNet, and transformer-based architecture~\cite{cho2021semantic}, denoted Ours w/CATs. We used 256x256 size for the input image and 16x16 size for the feature map. The learning rate is adjusted, starting from differently 3e-5 and 3e-6 for feature extraction and cost aggregation, respectively, and adjusted using AdamW optimizer. We set ${\alpha}_1={0.1}$, ${\alpha}_2={0.05}$, $\lambda^\mathcal{C} = 0.5$, and $\lambda^\mathcal{A} = 0.5$.

\subsection{Experimental Settings}
In this section, we demonstrate that our method is effective through comparison with others; WeakAlign~\cite{rocco2018end}, RTNs~\cite{kim2018recurrent}, NCNet~\cite{rocco2018neighbourhood}, DCCNet~\cite{huang2019dynamic}, 
ANCNet~\cite{li2020correspondence},
PMNC~\cite{lee2021patchmatch}, 
and CATs~\cite{cho2021semantic}. We also conduct an analysis of each component in our framework in the ablation study. To evaluate semantic matching, Proposal Flow~\cite{ham2017proposal} and TSS~\cite{taniai2016joint} benchmarks were used. The Proposal Flow benchmark contains PF-Pascal and PF-Willow~\cite{ham2017proposal}. TSS dataset is split into three subgroups: FG3D, JODS and PASCAL~\cite{taniai2016joint}. A percentage of correct keypoints (PCK) is employed for evaluation. 

\subsection{Experimental Results}
\begin{figure}[t]
	\centering
	\renewcommand{\thesubfigure}{}
	\subfigure[]
	{\includegraphics[width=0.195\linewidth]{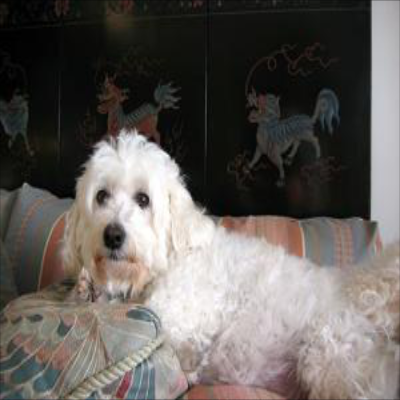}}\hfill
	\subfigure[]
	{\includegraphics[width=0.195\linewidth]{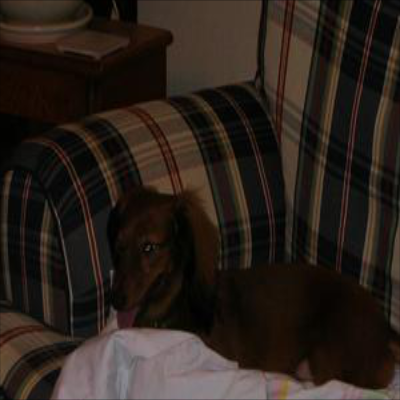}}\hfill
	\subfigure[]
	{\includegraphics[width=0.195\linewidth]{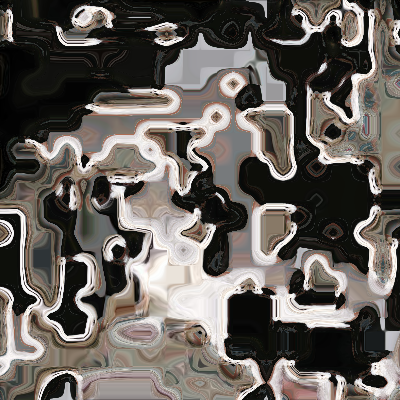}}\hfill
	\subfigure[]
	{\includegraphics[width=0.195\linewidth]{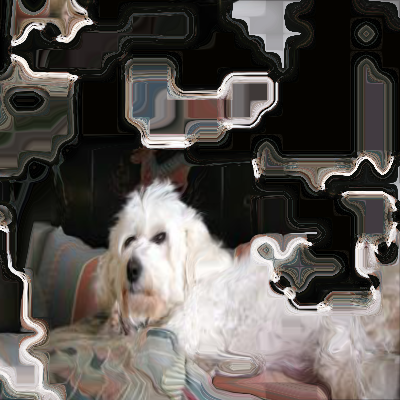}}\hfill
	\subfigure[]
	{\includegraphics[width=0.195\linewidth]{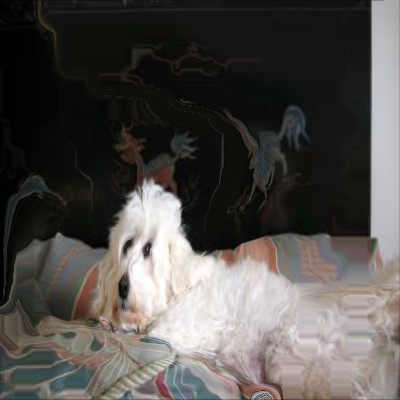}}\hfill \\
		\vspace{-21.5pt}
	\subfigure[]
	{\includegraphics[width=0.195\linewidth]{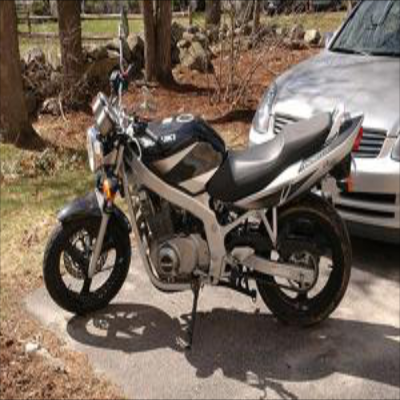}}\hfill
	\subfigure[]
	{\includegraphics[width=0.195\linewidth]{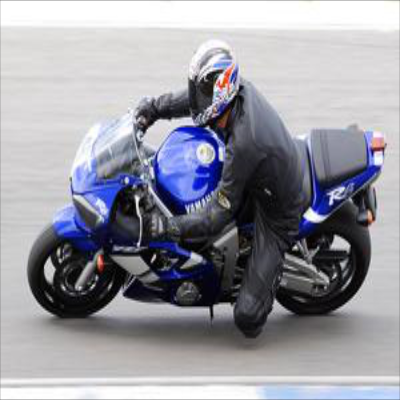}}\hfill
	\subfigure[]
	{\includegraphics[width=0.195\linewidth]{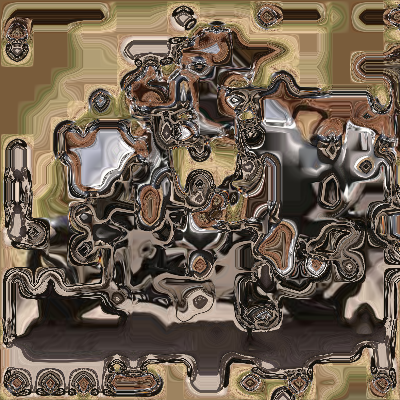}}\hfill
	\subfigure[]
	{\includegraphics[width=0.195\linewidth]{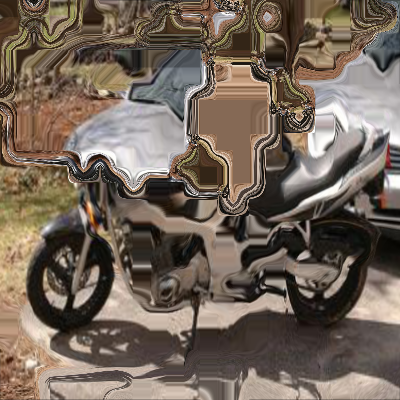}}\hfill
	\subfigure[]
	{\includegraphics[width=0.195\linewidth]{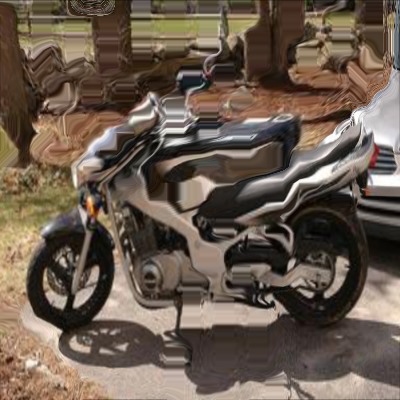}}\hfill \\
		\vspace{-21.5pt}
	{\includegraphics[width=0.195\linewidth]{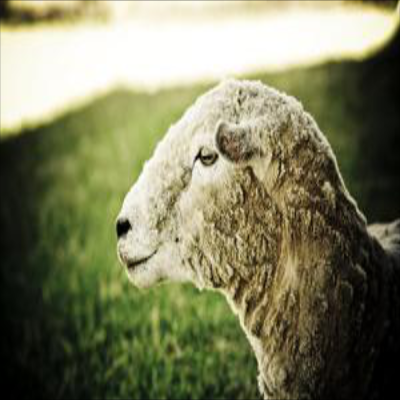}}\hfill
	\subfigure[]
	{\includegraphics[width=0.195\linewidth]{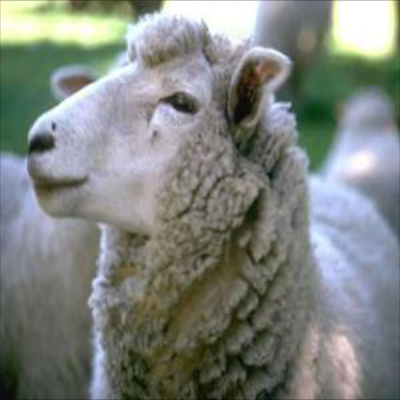}}\hfill
	\subfigure[]
	{\includegraphics[width=0.195\linewidth]{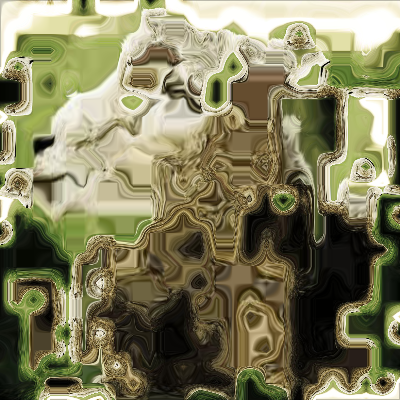}}\hfill
	\subfigure[]
	{\includegraphics[width=0.195\linewidth]{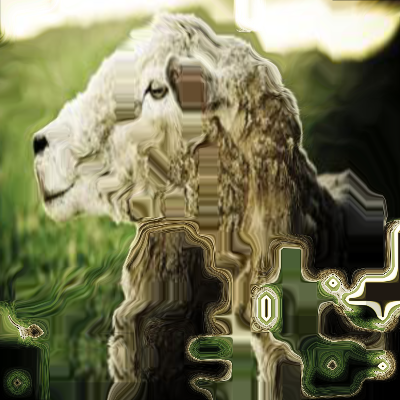}}\hfill
	\subfigure[]
	{\includegraphics[width=0.195\linewidth]{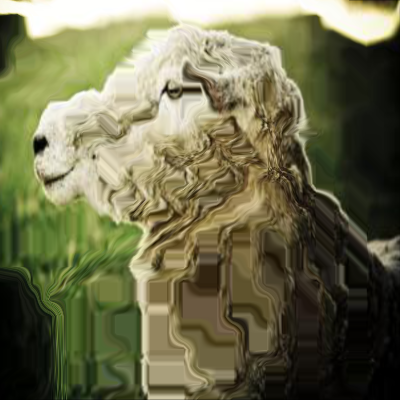}}\hfill \\
		\vspace{-21.5pt}

\subfigure[(a) $I_s$]
	{\includegraphics[width=0.195\linewidth]{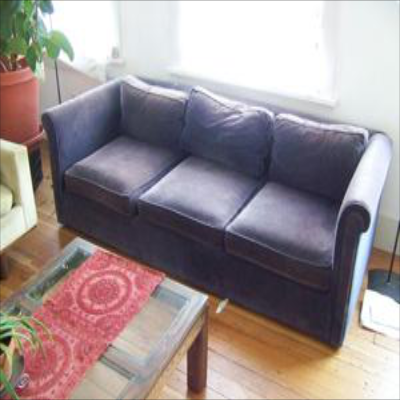}}\hfill
	\subfigure[(b) $I_t$]
	{\includegraphics[width=0.195\linewidth]{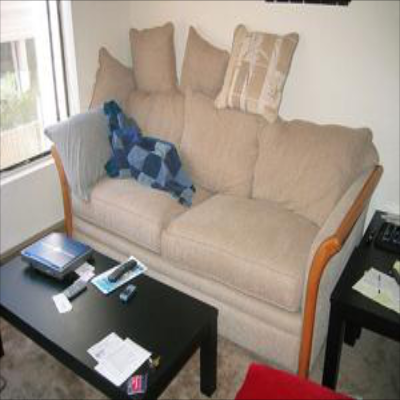}}\hfill
	\subfigure[(c) WTA]
	{\includegraphics[width=0.195\linewidth]{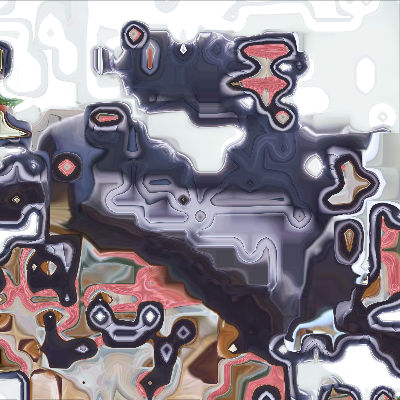}}\hfill
	\subfigure[(d) NCNet \cite{rocco2018neighbourhood}]
	{\includegraphics[width=0.195\linewidth]{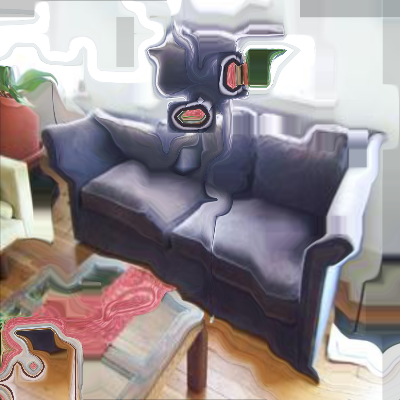}}\hfill
	\subfigure[(e) Ours]
	{\includegraphics[width=0.195\linewidth]{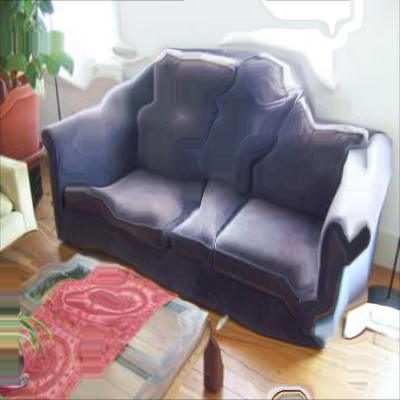}}\hfill \\
    \vspace{-10pt}
	\caption{\textbf{Qualitative results on PF-Pascal dataset~\cite{ham2017proposal}.}}	\vspace{-10pt}
	\label{fig:qual_network}
\end{figure}

~\tabref{tab:quan} shows the quantitative results. We conduct experiments on various benchmarks, such as PF-Pascal~\cite{ham2017proposal}, PF-Willow~\cite{ham2017proposal}, and TSS~\cite{taniai2016joint}. Our method records higher accuracy than both baselines, NCNet and CATs, by 1.1 and 5.2 on PF-Pascal, 2.3 and 2.9 on PF-Willow, respectively. Specifically, ours w/CATs outperforms other methods over all benchmarks and show the biggest performance improvement about 2 on PF-Pascal dataset compared to 90.6, which is the state-of-the-art result~\cite{lee2021patchmatch} among the similar network architectures and algorithms. Ours w/NCNet shows the highest performance on the FG3D as 95.0, and records average PCK of 78.4 on TSS benchmark. The qualitative results of semantic matching on the PF dataset are shown in~\figref{fig:qual_network}. (c), (d), (e) are warped images from (a) to (b) by WTA, NCNet~\cite{rocco2018neighbourhood}, and ours, respectively. Through (c), we could observe that errors of matches produced from feature extraction network affect the final output. Compared to (d), (e) shows accurate matching results even in difficult examples with occlusion, background clutter, and repetitive textures. 

\begin{table}[t]
\begin{center}
\begin{tabular}{cl|c}
\hlinewd{0.8pt}
\multicolumn{2}{c|}{Components}     & Accuracy \\ \hline
\textbf{(a)}   & Ours                    &   \textbf{80.0}        \\ \hline
\textbf{(b)}  & (-) Joint learning    &  78.4       \\
\textbf{(c)} & (-) Confidence-aware loss   &  77.7       \\ \hlinewd{0.8pt}
\end{tabular}
\end{center}
\vspace{-15pt}
\caption{\textbf{Ablation study on our modules.}}
\label{tab:ablation_1}
\end{table}

\begin{table}[t]
\begin{center}
\scalebox{1.0}{
\begin{tabular}{ccccc|c}
\hlinewd{0.8pt}
\multicolumn{5}{c|}{Loss component} & \multirow{2}{*}{PCK ($\alpha$ = 0.1)} \\ \cline{1-5}
 & $\mathcal{L}^{\mathcal{C}}_\mathrm{ccl}$ & $\mathcal{L}^{\mathcal{A} \rightarrow \mathcal{C}}_\mathrm{ccl}$ &$\mathcal{L}^{\mathcal{A}}_\mathrm{ccl}$& $\mathcal{L}^{\mathcal{C} \rightarrow \mathcal{A}}_\mathrm{ccl}$&                      \\ \hline
\textbf{(a)}&                     - &    -   &  \cmark   &   -     & 70.0                \\
\textbf{(b)}&                     - &    -   &  \cmark &\cmark & 71.7                \\
\textbf{(c)}&                \cmark &    -   & \cmark  &    -  & 78.4                \\
\textbf{(d)}&                \cmark &   \cmark&   \cmark&\cmark & \textbf{80.0}      \\ 
\hlinewd{0.8pt}
\end{tabular}
}
\vspace{-15pt}
\end{center}
\caption{\textbf{{Ablation study of our loss formulation.}}}
\label{tab:ablation_2}
\vspace{-10pt}
\end{table}

\subsection{Ablation Study}
In this section, we analyze the main components in our method, confidence-aware contrastive loss and joint learning, with NCNet baseline~\cite{rocco2018neighbourhood} on PF-Pascal. First, in ~\tabref{tab:ablation_1} we validate the effectiveness of joint learning and confidence-aware contrastive loss by the lower performance of \textbf{(b)} and \textbf{(c)} compared to \textbf{(a)} which has both of these components. This proves that training two networks in a complementary manner boosts the performance and confidence-aware contrastive loss leads the unreliable pseudo labels to be filtered out during the training process. 
We also verify the effectiveness of each confidence-aware contrastive loss component through possible combinations of components displayed in ~\tabref{tab:ablation_2}. Compared to \textbf{(a)} and \textbf{(b)}, both of which train two networks only with the loss from the aggregation module, \textbf{(c)} and \textbf{(d)} show better PCK results by using separate losses that come from feature extraction and cost aggregation respectively. From this, we can confirm that the direct loss from each module works as a sufficient supervision signal for training which is free from gradient vanishing. Based on the performance improvements observed between \textbf{(b)} and \textbf{(a)}, and between \textbf{(d)} and \textbf{(c)}, we can also confirm that the reliable sample from one module helps training the other module, as it supports the formulation of better loss signals. 

\section{Conclusion}
We address the limitations of the existing methods by jointly training feature extraction networks and aggregation networks in an end-to-end manner with the proposed confidence-aware contrastive loss. By jointly learning the networks with a novel loss function, our model outperforms the baseline and shows competitive results on standard benchmarks.

\noindent\textbf{Acknowledgements.}
This research was supported by the National Research Foundation of Korea (NRF-2021R1C1C1006897). 
\newpage
\bibliographystyle{IEEEbib}
\bibliography{strings,refs}

\begin{thebibliography}{10}

\bibitem{taira2019right}
Hajime Taira, Ignacio Rocco, Jiri Sedlar, Masatoshi Okutomi, Josef Sivic, Tomas
  Pajdla, Torsten Sattler, and Akihiko Torii,
\newblock ``Is this the right place? geometric-semantic pose verification for
  indoor visual localization,''
\newblock in {\em ICCV}, 2019.

\bibitem{wang2020learning}
Qianqian Wang, Xiaowei Zhou, Bharath Hariharan, and Noah Snavely,
\newblock ``Learning feature descriptors using camera pose supervision,''
\newblock in {\em ECCV}, 2020.

\bibitem{cho2021semantic}
Seokju Cho, Sunghwan Hong, Sangryul Jeon, Yunsung Lee, Kwanghoon Sohn, and
  Seungryong Kim,
\newblock ``Semantic correspondence with transformers,''
\newblock {\em arXiv:2106.02520}, 2021.

\bibitem{lee2019sfnet}
Junghyup Lee, Dohyung Kim, Jean Ponce, and Bumsub Ham,
\newblock ``Sfnet: Learning object-aware semantic correspondence,''
\newblock in {\em CVPR}, 2019.

\bibitem{rocco2018neighbourhood}
Ignacio Rocco, Mircea Cimpoi, Relja Arandjelovi{\'c}, Akihiko Torii, Tomas
  Pajdla, and Josef Sivic,
\newblock ``Neighbourhood consensus networks,''
\newblock {\em arXiv:1810.10510}, 2018.

\bibitem{lowe2004distinctive}
David~G Lowe,
\newblock ``Distinctive image features from scale-invariant keypoints,''
\newblock in {\em IJCV}, 2004.

\bibitem{choy2016universal}
Christopher~B Choy, JunYoung Gwak, Silvio Savarese, and Manmohan Chandraker,
\newblock ``Universal correspondence network,''
\newblock {\em arXiv:1606.03558}, 2016.

\bibitem{kim2017fcss}
Seungryong Kim, Dongbo Min, Bumsub Ham, Sangryul Jeon, Stephen Lin, and
  Kwanghoon Sohn,
\newblock ``Fcss: Fully convolutional self-similarity for dense semantic
  correspondence,''
\newblock in {\em CVPR}, 2017.

\bibitem{rocco2017convolutional}
Ignacio Rocco, Relja Arandjelovic, and Josef Sivic,
\newblock ``Convolutional neural network architecture for geometric matching,''
\newblock in {\em CVPR}, 2017.

\bibitem{rocco2018end}
Ignacio Rocco, Relja Arandjelovi{\'c}, and Josef Sivic,
\newblock ``End-to-end weakly-supervised semantic alignment,''
\newblock in {\em CVPR}, 2018.

\bibitem{melekhov2019dgc}
Iaroslav Melekhov, Aleksei Tiulpin, Torsten Sattler, Marc Pollefeys, Esa Rahtu,
  and Juho Kannala,
\newblock ``Dgc-net: Dense geometric correspondence network,''
\newblock in {\em WACV}, 2019.

\bibitem{rocco2020efficient}
Ignacio Rocco, Relja Arandjelovi{\'c}, and Josef Sivic,
\newblock ``Efficient neighbourhood consensus networks via submanifold sparse
  convolutions,''
\newblock in {\em ECCV}, 2020.

\bibitem{li2020correspondence}
Shuda Li, Kai Han, Theo~W Costain, Henry Howard-Jenkins, and Victor Prisacariu,
\newblock ``Correspondence networks with adaptive neighbourhood consensus,''
\newblock in {\em CVPR}, 2020.

\bibitem{lee2021patchmatch}
Jae~Yong Lee, Joseph DeGol, Victor Fragoso, and Sudipta~N Sinha,
\newblock ``Patchmatch-based neighborhood consensus for semantic
  correspondence,''
\newblock in {\em CVPR}, 2021.

\bibitem{leibe2008robust}
Bastian Leibe, Ale{\v{s}} Leonardis, and Bernt Schiele,
\newblock ``Robust object detection with interleaved categorization and
  segmentation,''
\newblock in {\em IJCV}, 2008.

\bibitem{park2020contrastive}
Taesung Park, Alexei~A Efros, Richard Zhang, and Jun-Yan Zhu,
\newblock ``Contrastive learning for unpaired image-to-image translation,''
\newblock in {\em ECCV}, 2020.

\bibitem{wang2020dense}
Xinlong Wang, Rufeng Zhang, Chunhua Shen, Tao Kong, and Lei Li,
\newblock ``Dense contrastive learning for self-supervised visual
  pre-training,''
\newblock {\em arXiv:2011.09157}, 2020.

\bibitem{ham2017proposal}
Bumsub Ham, Minsu Cho, Cordelia Schmid, and Jean Ponce,
\newblock ``Proposal flow: Semantic correspondences from object proposals,''
\newblock in {\em TPAMI}, 2017.

\bibitem{taniai2016joint}
Tatsunori Taniai, Sudipta~N Sinha, and Yoichi Sato,
\newblock ``Joint recovery of dense correspondence and cosegmentation in two
  images,''
\newblock in {\em CVPR}, 2016.

\bibitem{sundaram2010dense}
Narayanan Sundaram, Thomas Brox, and Kurt Keutzer,
\newblock ``Dense point trajectories by gpu-accelerated large displacement
  optical flow,''
\newblock in {\em ECCV}, 2010.

\bibitem{meister2018unflow}
Simon Meister, Junhwa Hur, and Stefan Roth,
\newblock ``Unflow: Unsupervised learning of optical flow with a bidirectional
  census loss,''
\newblock in {\em AAAI}, 2018.

\bibitem{liu2019ddflow}
Pengpeng Liu, Irwin King, Michael~R Lyu, and Jia Xu,
\newblock ``Ddflow: Learning optical flow with unlabeled data distillation,''
\newblock in {\em AAAI}, 2019.

\bibitem{kim2018recurrent}
Seungryong Kim, Stephen Lin, Sang~Ryul Jeon, Dongbo Min, and Kwanghoon Sohn,
\newblock ``Recurrent transformer networks for semantic correspondence,''
\newblock in {\em NeurIPS}, 2018.

\bibitem{huang2019dynamic}
Shuaiyi Huang, Qiuyue Wang, Songyang Zhang, Shipeng Yan, and Xuming He,
\newblock ``Dynamic context correspondence network for semantic alignment,''
\newblock in {\em ICCV}, 2019.

\bibitem{he2016deep}
Kaiming He, Xiangyu Zhang, Shaoqing Ren, and Jian Sun,
\newblock ``Deep residual learning for image recognition,''
\newblock in {\em CVPR}, 2016.

\bibitem{deng2009imagenet}
Jia Deng, Wei Dong, Richard Socher, Li-Jia Li, Kai Li, and Li~Fei-Fei,
\newblock ``Imagenet: A large-scale hierarchical image database,''
\newblock in {\em CVPR}, 2009.

\end{thebibliography}

\end{document}